# Seeing eye-to-eye?
# A comparison of object recognition performance in humans and deep convolutional neural networks under image manipulation

Leonard Elia van Dyck[1,2*] & Walter Roland Gruber[1,2]

[1]*University of Salzburg, Department of Psychology, Austria*
[2]*University of Salzburg, Center for Cognitive Neuroscience, Austria*
*\*To whom correspondence should be addressed:* leonard.vandyck@sbg.ac.at

**Abstract**

For a considerable time, deep convolutional neural networks (DCNNs) have reached human benchmark performance in object recognition. On that account, computational neuroscience and the field of machine learning have started to attribute numerous similarities and differences to artificial and biological vision. This study aims towards a behavioral comparison of visual core object recognition performance between humans and feedforward neural networks in a classification learning paradigm on an ImageNet data set. For this purpose, human participants (n = 65) competed in an online experiment against different feedforward DCNNs. The designed approach based on a typical learning process of seven different monkey categories included a training and validation phase with natural examples, as well as a testing phase with novel, unexperienced shape and color manipulations. Analyses of accuracy revealed that humans not only outperform DCNNs on all conditions, but also display significantly greater robustness towards shape and most notably color alterations. Furthermore, a precise examination of behavioral patterns highlights these findings by revealing independent classification errors between the groups. The obtained results show that humans contrast strongly with artificial feedforward architectures when it comes to visual core object recognition of manipulated images. In general, these findings are in line with a growing body of literature, that hints towards recurrence as a crucial factor for adequate generalization abilities.

# 1 Introduction

Similar to other fields of artificial intelligence, technological progress in computer vision has initiated the highly philosophical discussion about a possible successor of the human brain. Artificial vision is attracting widespread and interdisciplinary interest, as for some it is believed to provide a revolutionary step towards intelligent machines in allowing them to perceive, process, and interact with the external world [1]. Primates including humans have mastered a skill crucial to higher-level cognition. It is one key ability of visual perception to succeed at labeling a large number of objects, most often independent of latent variables such as position, size, pose, or illumination [2, 3, 4], that has awarded some species numerous evolutionary advantages over others. In a way, view-invariant core object recognition serves as a foundation for many learning processes, where classification of chaotic incoming visual stimuli into meaningful constellations forms retrievable categories [5]. For many years, neuroscience has attributed this skill to a chain of visual perception areas from retinal ganglion cells (RGC), to lateral geniculate nucleus (LGN), and ventral visual pathway, which traverses visual areas V1, V2, and V4, in order to end in inferior temporal cortex (ITC) [6, 7, 8, 9]. However, in the last few years, upcoming feedforward deep convolutional neural networks (DCNNs) have been used to emulate this process as a result of matching performance and attested representational similarities [5, 10, 11, 12, 13]. To this day, mechanisms of core object recognition in the primate brain have not been fully uncovered. However, computer science has managed to engineer biology-inspired artificial replicas, which currently serve as state-of-the-art models, without a more extensive understanding of the underlying coherences [14].

In line with earliest pioneers of machine intelligence such as Alan Turing, performance optimization towards defined benchmarks is still cherished in the field. Today, both biological and artificial information processors undoubtably compete on the same level in organized events such as the ImageNet Large Scale Visual Recognition Challenge (ILSVRC) [15]. An arms race of modeling *in silico* vision has resulted in accuracies undreamt-of. The development can be demonstrated by the winners of milestone events such as AlexNet in 2012 [16], GoogLeNet in 2014 [17], and ResNet in 2015 [18] with corresponding top-5 error rates of 18.9 % up to 3.57 %. DCNNs have since proven themselves repeatedly to achieve respectable results in image recognition tasks [1, 19]. Similar to primate visual areas, feedforward multilayer neural networks are hierarchically structured with many layers of nodes, the technical abstraction models of biological neurons. While the artificial *backpropagation* learning algorithm is widely considered as biologically implausible [21], other simple operations performed in individual layers appear, at least theoretically, implementable in biological circuits [22].

Although, up to date DCNNs are capable of outperforming their human prototypes in recognizing natural images, when trained on sufficient data [16, 18, 23, 24, 25], these same systems fail at more abstract and atypical examples, which do not pose a problem to their creators. Unlike primates, feedforward neural networks are extremely sensitive to distribution shifts, such as random noise or blur [26, 27, 28], and can be fooled by a simple change of background [29, 30], object rotation [31], object texture [32], or by shift of a few pixels only [33]. Nevertheless, an extensive body of literature suggests that many of these phenomena are not the result of lacking generalization abilities [28, 34, 35, 36], but rather a consequence of artificial systems falling for unintended, nonhuman shortcut learning strategies, which in turn lead to undesirable results [14].

As the field of neuroscience uses DCNNs as a scientific multitool for prediction, explanation, and exploration simultaneously [22], a heated debate about the coherence of primate brain areas and artificial neural networks in visual perception has emerged. Findings suggest that both processing systems may share similar ways of representing visual information, as early and late layers of neural networks predict low- and high-level visual brain areas, respectively [37]. Moreover, better performing neural networks were found to be more similar to ITC with more clustered between-category representation patterns [38, 39, 40, 41], as well as greater within-category dissimilarities. Here, another key aspect is supervised learning, as neural networks require an enormous amount of labeled data in order to acquire robust representations. Through the process of providing neural networks with important categorical distinctions found in biological vision, such as animate versus inanimate, and faces versus objects, labeling accuracy and explanation of ITC data were both found to increase [10].

In contrast, recent studies propose representational differences of visual information as only early layers of neural networks seem to capture the structure of associated lower visual areas [42]. Previous work of Kar et al. (2019) [43] compared performance of humans, monkeys, and well-studied AlexNet in a core object recognition task using electrophysiologically assigned *early-* and *late-solved* images. Here, the population of emerging activity in monkeys' ITC was noticeably slower (~ 30 ms) for *late-solved* challenge images, agreeing with the assumption that neural timing in mid- and high-level visual brain areas operates in a most non-linear fashion [44]. Additionally, performance results show that a highly recurrent primate ventral stream outperforms a strictly feedforward DCNN predominantly at these *late-solved* challenge images. As can be seen here, a considerable amount of literature suggests that in addition to simple feedforward mechanisms, challenging images might need extra processing steps performed by recurrent circuits [45] with neurons interconnected to loops [46]. Since feedforward neural networks are missing recurrent connectivity as naturally occurring in the primate visual system [47, 48, 49], these networks should have no memory and therefore exhibit serious problems with previously unexperienced shape and color manipulation of natural images.

Therefore, the presented work aims towards testing hypotheses generated in accordance with literature on image recognition performance of human and computer vision under commonly studied manipulations. However, it should be clarified, that the study at-hand does not represent an attempt to compare solely biological and artificial feedforward architectures, meaning that underlying computations are disregarded, and human vision is assumed to include both feedforward and recurrent processes. As shown by numerous examinations DCNNs should perform indistinguishable on natural images when compared to human observers. However, their accuracy for shape and color manipulated challenge images should be worse due to missing recurrent computations needed to classify these predominantly *late-solved* examples. Additionally, humans are thought to display great robustness towards color alterations [50], which could be regarded as a key feature of astonishing core object recognition abilities, but also as a result of the higher order processes of abstraction and imagination. Consequently, behavioral classification patterns between recurrent *in vivo* systems and non-recurrent *in silico* replicas are assumed to diverge with uncorrelated classification errors.

Based on fundamental principles of comparative psychology, this investigation tries to achieve increased species-fairness in the performed human-machine comparisons by adopting ideas suggested by Firestone (2020) [51]. Even though completely fair comparisons might not be realizable, several measures were taken in order to limit machines like humans and humans like machines.

# 2 Materials and Methods

## 2.1 Data Set

In order to draw a behavioral comparison in performance of human participants and three feedforward DCNNs AlexNet, GoogLeNet, and ResNet-50 on the same specific, novel task, a general classification learning paradigm including a training, validation, and testing phase was adopted. Presented images were all part of a small subset included in the ImageNet database consisting of seven different species of monkeys, apes, and lemurs. An approach regarding both sufficient variance as well as participants' attentional and motivational capabilities was decided as most appropriate for this investigation. Consequently, less categories with higher similarity and in turn higher difficulty were selected. The initial sample consisted of 30 labeled and randomly drawn images for each of the seven subsets with their exact designations *Gorilla, Gorilla gorilla*; *Chimpanzee, chimp, Pan troglodytes*; *Orangutan, orang, orangutan, Pongo pygmaeus*; *Gibbon, Hylobates lar*; *Spider monkey, Ateles geoffroyi*; *Baboon*; and *Madagascar cat, ring-tailed lemur, Lemur catta*. As the study was originally conducted in German, presented labels were translated to their respective, conventional term. Of the 30 images per class, 15 were used for training, 3 for validation, and 12 for testing purposes. Hence, the data set consisted of 210 images in total, splitting up to 105 training images, 21 validation images, and 84 testing images. Moreover, challenge images for testing were randomly assigned to three increasing levels of shape distortion and a color alteration with three primary color variations, while keeping an even distribution of manipulations within categories (see Figure 1). All images were preprocessed using the image manipulation program GIMP (Version 2.10, The GIMP Development Team) in a standardized way by centering towards the dominant object if pixel size exceeded the demands, as well as thoroughly checking for and eliminating examples with multiple dominant objects and replicas (e.g., drawing or statue). Finally, challenge images were compiled using a shape manipulation called *whirl effect* at the three intensity levels 50 %, 150 %, and 250 % (hereafter *shape50*, *shape150*, and *shape250*) distorting the image in a concentric way, and a color manipulation overlaying 100 % intensity of random and equally distributed primary colors red, green, and blue (hereafter *color*) to the entire image [52].

In this way, both groups of human participants and DCNNs were presented the exact same images in each of the experiment phases with a balanced distribution of category examples and manipulated variations. The *whirl effect* was chosen as a form of shape manipulation in order to rule out any evolutionary and/or training advantages, as neither humans nor DCNNs should have experienced it before, while for color alterations primary colors were applied, as both biological and artificial vision systems are based on their combination. Adversarial examples were considered to be unsuitable for this comparison, as they are generally designed to fool artificial neural networks but not humans.

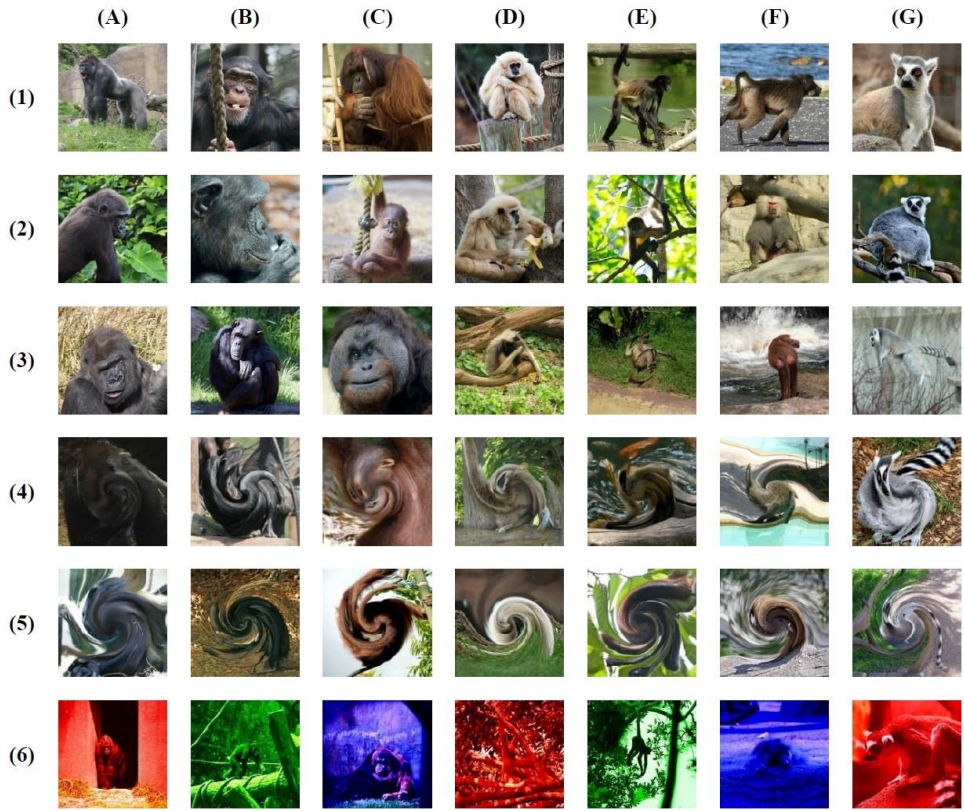

**Figure 1.** Examples of the obtained ImageNet data set. Letters show chosen ImageNet subsets (A: *gorilla*, B: *chimpanzee*, C: *orangutan*, D: *gibbon*, E: *spider monkey*, F: *baboon*, G: *ring-tailed lemur*) and numbers show levels of manipulation (1: *training/natural*, 2: *validation/natural*, 3: *testing/shape50*, 4: *testing/shape150*, 5: *testing/shape250*, 6: *testing/color*).

## 2.2 Participants and Deep Neural Networks

### 2.2.1 Human Participants

A total of 65 human participants were recruited for the online experiment *Object recognition – biological and artificial neural networks*. The representative sample consisted of 34 individuals identifying as women and 31 as men. The anonymous observers were between the ages of 19 and 65 (M = 30.49; SD = 13.95) and reported normal or corrected-to-normal vision without any problems of color perception. Participants had to complete the study on a computer with a keyboard and were advised to do so in a quiet and undisturbed environment. Instructions informed about the importance of adequate brightness and avoidance of reflecting light on the screen. Participation was rewarded with 0.5 accredited study completion hours for the 31 psychology students. All procedures were in alignment with the declaration of Helsinki and agreed to by participants with informed consent.

### 2.2.2 Human Experiments

In a 15-minute online study, participants had to complete all three stages of a typical classification process with images of the data set in randomized order. First, in the training phase, 15 consecutive sample images of each of the seven classes were presented for 1500 ms on a blank white screen with respective labels shown as a caption. In-between trials, subjects had to focus on a fixation cross for 500 ms. After individual category sets, in order to advance, participants had to press a button. This sequence was implemented to improve compliance with the instructions. Second, in the validation phase, presented natural images had to be labeled by pressing a corresponding number on the keyboard. Adequate feedback including the correct answer was provided for 1500 ms. Participants had to answer in a forced-choice format without a provided *other* category as a measure to limit humans like machines and therefore provide a fair comparison with DCNNs that do not possess such an option. Furthermore, this constraint allows for equal *guessing* percentages between both groups. Last, in the testing phase, this process was continued for manipulation images without any feedback (see Figure 2). Classification decisions in both validation and testing phase were self-paced and participants were informed by the instructions to complete the task according to their individual speed. This approach was chosen over a rapid classification task with set time limits in order to ensure careful conduction without confounding reaction time aspects. As a result, performance data for a total of 6825 trials were recorded. The online study was implemented in German and programmed using the open source graphical experiment builder OpenSesame (Version 3.2.8, OpenSesame, Amsterdam, The Netherlands) [53].

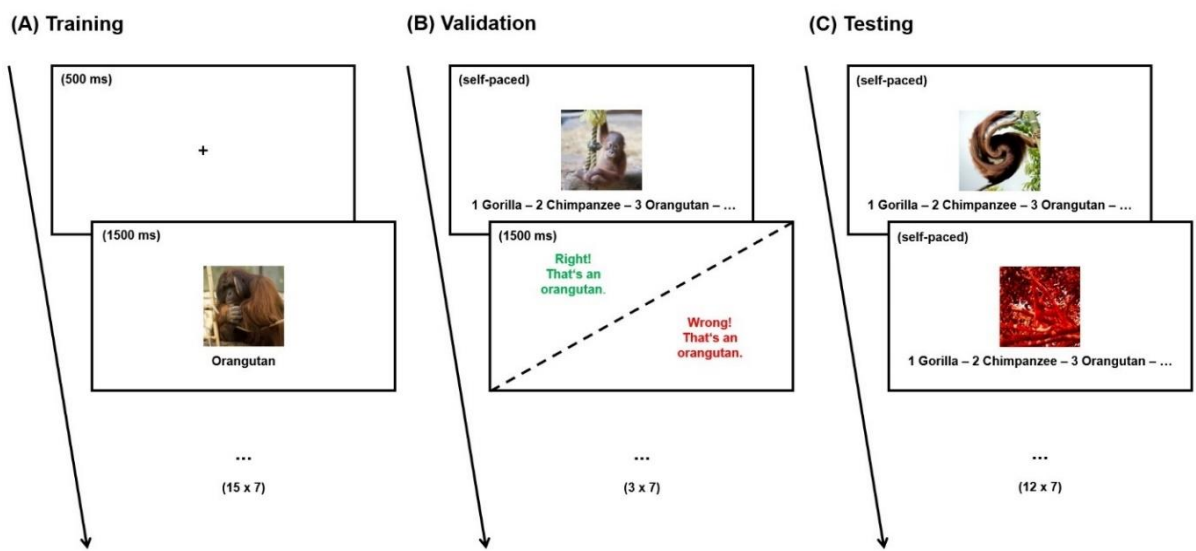

**Figure 2.** Experimental setup of the online study. At first, in the training phase (A), participants were shown 15 control images per category for 1500 ms with a fixation cross in-between trials for 500 ms. Then, in the validation phase (B), participants classified 3 control images per category following individual feedback for 1500 ms. Finally, in the testing phase (C), participants continued labelling 12 challenge images per category in randomized order.

### 2.2.3 Deep Neural Networks

To investigate DCNN performance, the three neural networks AlexNet, GoogLeNet, and ResNet-50 were trained and tested entirely in MATLAB with Deep Learning Toolbox (Version 9.8, R2020a, The MathWorks Inc., Natick, MA, USA). All networks were acquired as pretrained models of the ILSVRC data subset with more than 1.2 million training images, 50000 validation images, 150000 testing images, and 1000 classes [16]. In a fine-tuning process, the last layers of the networks were replaced to reduce classification to the seven classes preexistent in the data. Subsequently, the models were trained on clean control images with a smaller learning rate of 0.0001 to conserve weights in early and solely adjust in later layers. Training parameters were standardized for all networks and the process was stopped as soon as the loss on the validation set did not decrease further than a pre-set accepted error. While the chosen networks share most of their architecture, they can be characterized by a few individual structural differences. In comparison to 8-layer AlexNet, 22-layer GoogLeNet benefits from so called *inception modules* which drastically reduce the number of parameters [17], and 50-layer ResNet-50 uses residual learning with *skip connections* inspired by pyramidal cells in the cerebral cortex [18].

## 2.3 Statistical Analysis

Statistical analyses were performed with IBM SPSS Statistics (Version 26, IBM Corp., Chicago, IL, USA) and R (Version 3.6.1, R Core Team, Vienna, Austria) using the package ggplot2 [54] for data visualization. As several tests (Kolmogorov-Smirnov, visual inspection of distributions) revealed that performance data were not normally distributed ($p < .05$), in an attempt to replicate published findings under novel manipulations, several non-parametric tests were conducted. To control for outliers, two participants with results two standard deviations below mean performance (64.28 %) were dropped from further analysis. One-sample Wilcoxon signed-rank tests with r as effect size [55] were performed to investigate between-group differences comparing the human observer sample to each one of the three tested neural networks for all stages of manipulation, respectively. For main analyses, this method was chosen over the idea to treat artificial neural networks as a population of their own to explore possible differences based on architecture. Later on, for reasons of clarity and comprehensibility, DCNNs were treated as a group allowing statistically equal comparisons. Within-group differences for both manipulation types were explored using Friedman's ANOVA with Kendall's W as effect size followed by multiple Bonferroni-corrected pairwise comparisons. Furthermore, in order to contrast typical human and DCNN classification behavior across manipulations, confusion matrices of true and predicted labels were computed based on groups' average performances. Additionally, special confusion difference matrices were calculated to highlight striking distinctions in classification behavior, and similarity of errors by humans and neural networks was investigated using Spearman's rank order correlation with Spearman's $r_s$ as effect size. Aside from that, pretrained networks without a fine-tuning process were analyzed to guarantee a suitable methodological approach for comparability of the presented learning paradigm.

# 3 Results

## 3.1 Accuracy for Control and Challenge Images

As anticipated, even on a rather underpowered training data set of only few instead of many hundred newly learned examples per category, fine-tuned DCNNs achieve respectable results on natural images. Nevertheless, in the present study design all tested neural networks clearly miss performance statistically comparable to human observers (see Table 1). Remarkably, top-performing GoogLeNet reaches an accuracy of 80.95 % outperforming both AlexNet and ResNet-50 with 76.19 %, respectively. Generally speaking, our results show that the novel manipulations undoubtably affect recognition accuracy and lead to performance drops in both groups alike. However, as apparent in Figure 3, human participants ($Mdn_{human}$ = 73.81) outperform all three fine-tuned neural networks ($Mdn_{AlexNet}$ = 47.62; $Mdn_{GoogLeNet}$ = 52.38; $Mdn_{ResNet-50}$ = 46.43) by a significant margin at all levels of manipulation.

**Table 1.** Image recognition performance under different conditions. One-sample Wilcoxon signed-rank tests comparing individual neural networks to human participants. Median accuracy across levels with interquartile ranges (IQR) for human sample and fixed values for neural networks. Test statistic W and standardized test statistic z are reported with a significant difference at the .05 level (*) and at the .001 level (**).

|  |  | natural | manipulated | shape50 | shape150 | shape250 | color |
|---|---|---|---|---|---|---|---|
| humans | Mdn | 85.71 | 73.81 | 80.95 | 76.19 | 61.90 | 80.95 |
|  | interquartile range | 19.05 | 17.09 | 14.29 | 14.28 | 14.29 | 19.04 |
| AlexNet | Mdn | 76.19 | 47.62 | 66.67 | 52.38 | 33.33 | 38.10 |
|  | W | 1489.00 | 2013.00 | 1648.00 | 1888.00 | 2016.00 | 2016.00 |
|  | z | 5.66** | 6.88** | 5.41** | 6.79** | 6.92** | 6.92** |
|  | p-value | < .001 | < .001 | < .001 | < .001 | < .001 | < .001 |
| GoogLeNet | Mdn | 80.95 | 52.38 | 76.19 | 52.38 | 28.57 | 52.38 |
|  | W | 1028.00 | 2002.00 | 958.00 | 1888.00 | 2016.00 | 1949.00 |
|  | z | 2.78* | 6.81** | 2.16* | 6.79** | 6.92** | 6.83** |
|  | p-value | .005 | <.001 | .031 | <.001 | <.001 | <.001 |
| ResNet-50 | Mdn | 76.19 | 46.43 | 71.43 | 61.90 | 28.57 | 23.81 |
|  | W | 1489.00 | 1952.00 | 1440.00 | 1655.00 | 2016.00 | 2016.00 |
|  | z | 5.66** | 6.84** | 3.88** | 6.21** | 6.92** | 6.92** |
|  | p-value | < .001 | < .001 | < .001 | < .001 | < .001 | < .001 |

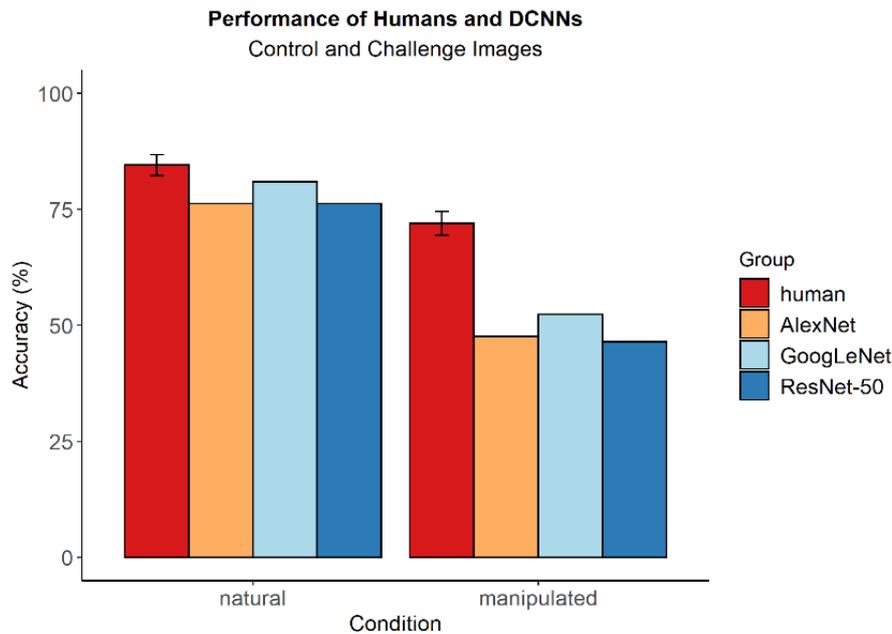

**Figure 3.** Performance of humans and DCNNs on control and challenge images. Human observers outperform tested DCNNs on both control and challenge images by a significant margin. Performance for challenge images is averaged over levels of shape and color manipulation. Error bars indicate 95 %-confidence intervals for human participants only, as DCNNs show fixed values without any variance.

### 3.1.1 Shape

With increasing distortion of shape properties, recognition ability decreases significantly for human participants ($\chi^2(3) = 122.65$, $p < .001$, $W = .65$). As additional post-hoc tests for human performance highlight, the effect is significant between all levels of intensity. Interestingly, these results can also be observed in fine-tuned neural networks, as their performance gradually decreases over levels of increasing distortion ($\chi^2(3) = 9.00$, $p < .029$, $W = 1$) (see Figure 4 A). In both groups statistical tests indicate large effects of shape manipulation. However, in contradiction to human performance, accuracy of neural networks is statistically indistinguishable between individual consecutive levels (see Table 2). As an exceptional case, GoogLeNet (Mdn = 76.19) accomplishes surprisingly solid performance on shape50 images in comparison to human benchmark capability (Mdn = 80.95) although this difference still reaches a significant level ($W = 958.00$, $z = 2.16$, $p = .031$, $n = 63$, $r = .27$).

**Table 2.** Multiple pairwise comparisons between individual levels of manipulation. Standardized test statistic z is shown. Median accuracy is significantly different at the .05 level (*) and at the .001 level (**). Dropped significant p-values after Bonferroni correction are indicated ($^\dagger$).

|        |          | natural | shape50       | shape150      | shape250       |
|--------|----------|---------|---------------|---------------|----------------|
| humans | natural  | 1       | 2.52<br>.012$^\dagger$ | 5.28**<br>< .001 | 10.28**<br>< .001 |
|        | shape50  |         | 1             | 2.76*<br>.035 | 7.76**<br>< .001 |
|        | shape150 |         |               | 1             | 5.00**<br>< .001 |
|        | shape250 |         |               |               | 1              |
| DCNNs  | natural  | 1       | .95<br>.343   | 1.90<br>.058  | 2.85*<br>.004  |
|        | shape50  |         | 1             | .95<br>.343   | 1.90<br>.058   |
|        | shape150 |         |               | 1             | .95<br>.343    |
|        | shape250 |         |               |               | 1              |

### 3.1.2 Color

Further examination suggests that in comparison to control images ($Mdn_{human} = 85.71$; $Mdn_{DCNN} = 76.19$) challenge images with an overlaying primary color ($Mdn_{human} = 80.95$; $Mdn_{DCNN} = 38.10$) are found to be recognized with significantly lower accuracy in human observers ($Ws = 121.00$, $z = -5.18$, $p < .001$, $n = 63$, $r = -.65$), while the group of tested neural networks did not show a statistically significant decrease ($Ws = 3.00$, $z = -1.60$, $p = .109$, $n = 3$, $r = -.93$). However, this should be a consequence of high intergroup variance between architectures. Again, statistical tests display large effects of color manipulation in both groups. Nonetheless, in line with our hypotheses, human subjects share strikingly greater robustness towards color manipulation, as their accuracy only decreases by 8.09 % while neural networks' performance seems to be heavily affected in a sharp decline of 39.68 % (see Figure 4 B).

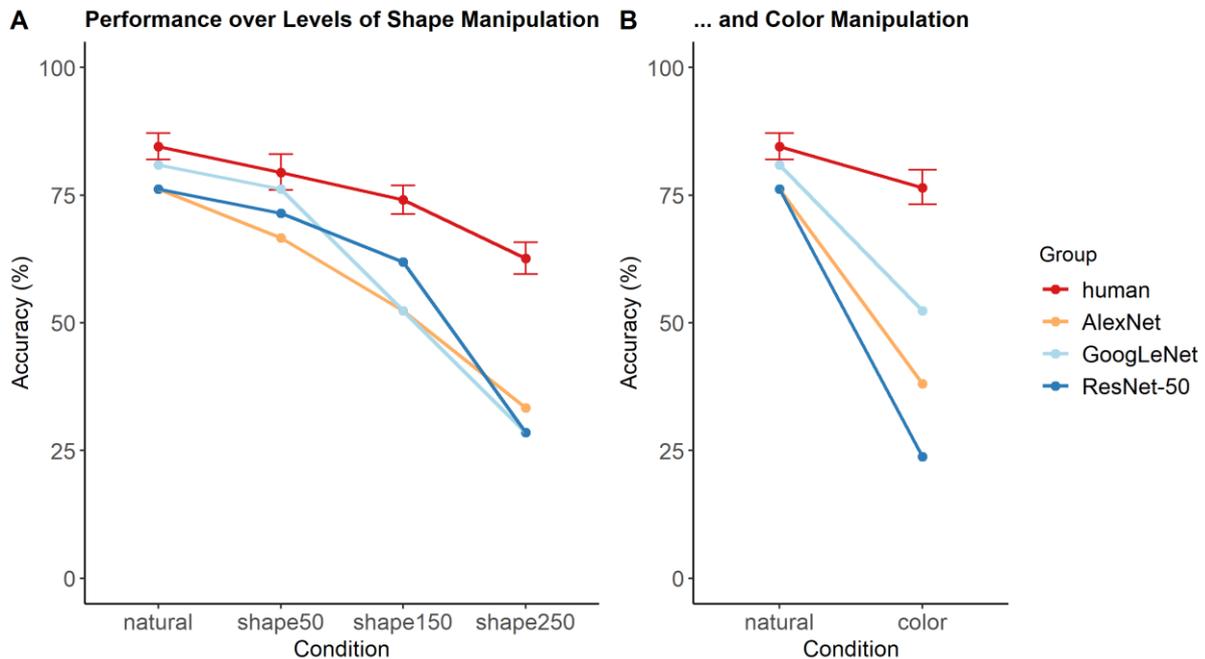

**Figure 4.** Performance of humans and DCNNs over levels of manipulation. Human observers outperform DCNNs across all levels of shape manipulation (A) and color manipulation (B) by a significant margin. Both groups experience significant recognition difficulties with increasing intensity of manipulation. Error bars indicate 95 %-confidence intervals for human participants only, as DCNNs show fixed values without any variance.

## 3.2 Confusion

As a way to explore behavioral classification patterns of the two groups for individual conditions in-depth, confusion matrices were computed. These visualizations are commonly used in the field of machine learning to examine possible issues underlying misclassification. Typically, as in the present case, true labels of images are plotted against predicted labels assigned by the observer, whereby a visualization with diagonally correct and off-diagonally incorrect responses becomes apparent. As indicated by Figure 5, average human responses (A-D) show a diagonally correct classification pattern slightly dissolving across shape distortion but staying nearly untouched by color manipulation. In addition, details suggest that observers occasionally seemed to confuse the categories *gibbon* and *spider monkey*. In contradiction to their human prototypes, DCNNs (E-H) display much more confusion across shape manipulation as the diagonal dissolves rapidly. Here, it is clearly visible that even color manipulation severely disrupts recognition ability in feedforward neural networks as the matrix illustrates poor performance with lots of confusion and accuracy close to chance level. As this paper focuses on the comparative aspects, further steps proceed very much in the same way as work by Geirhos et al. (2017) [56] with calculation of confusion difference matrices. This method is based on visualizing the difference between human and neural network behavioral patterns as the groups' mean performances are subtracted in a way to result in positive human (red) and negative DCNN (blue) occurrence entries. Surprisingly, as displayed in Figure 6, even though neural networks performed significantly worse on challenge images across all individual shape manipulations, they still seem to classify a few categories better than human observers under heavy distortion (see i.e., *gorilla* and *baboon*).

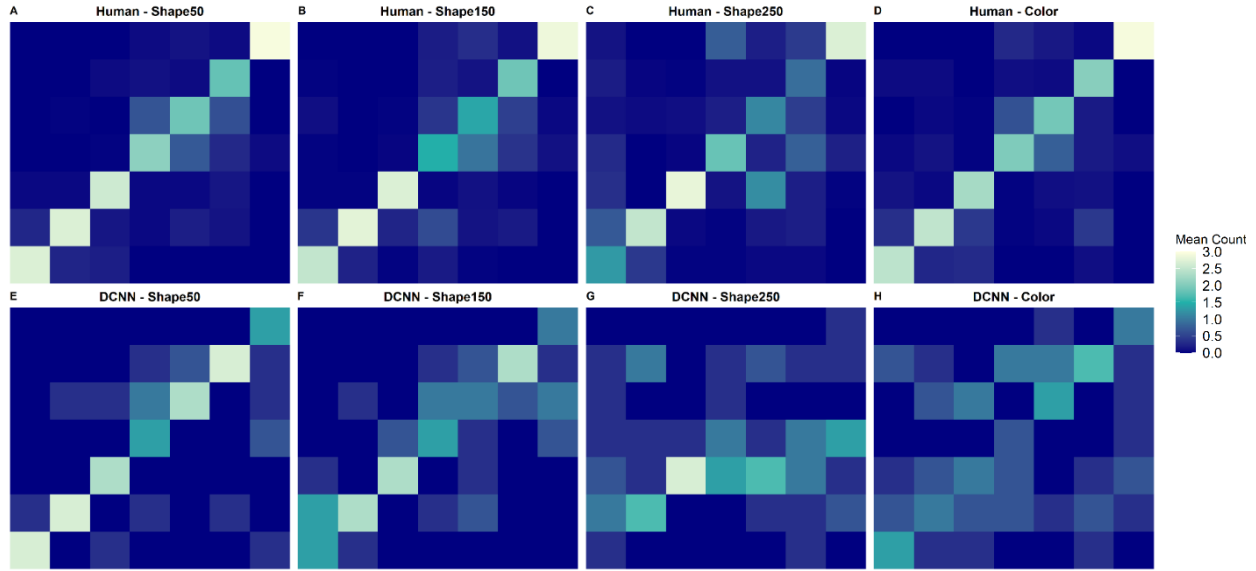

**Figure 5.** Confusion matrices for humans and DCNNs. Computed matrices show mean counts for predicted labels plotted against true labels with diagonally correct and off-diagonally incorrect entries. Classification patterns of human participants (A-D) reveal slightly increasing confusion across shape distortions and still great robustness towards color manipulations. In contrast, DCNNs (E-H) experience heavy confusion across shape and particularly color alterations.

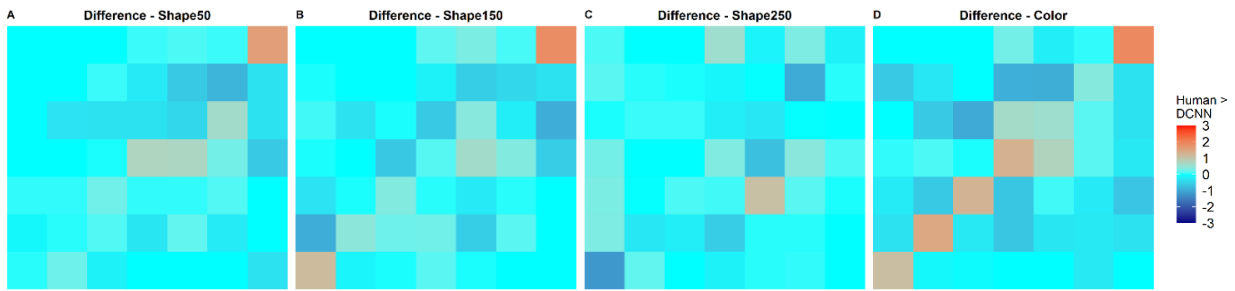

**Figure 6.** Confusion difference matrices for human versus DCNN occurrence. Computed matrices show classification occurrence on combinations of predicted and true labels. While positive (red) entries indicate preferred choice by humans, negative (blue) entries denote preferred choice by DCNNs. Difference matrices for shape distortions (A-C) suggest that neural networks generally experienced more misclassifications as most blue entries can be found off-diagonally. However, in some categories DCNNs still outperform human participants. The difference matrix for color alterations (D) clearly highlights a human predominance on images with a color overlay.

## 3.3 Error Correlation

Further statistical procedure confirms divergent misclassification patterns as several Spearman's rank correlations offer no compelling evidence for significantly correlated errors made by humans and DCNNs in most manipulations (see Table 3). Interestingly, against our expectations, correlation of committed classification errors suggest no significant coherence between AlexNet and the two other tested feedforward neural networks. However, careful attention must be exercised in interpreting these results, as the computed correlations may not be reliable for lower levels of distortion due to a small number of committed errors.

**Table 3.** Spearman's rank correlation coefficients $r_s$ indicate that misclassification patterns between human participants and individual or averaged DCNNs do not exhibit significant correlations for most manipulations. Averaged misclassifications ($^\dagger$). Correlations are significant at the .05 level (*) and at the .001 level (**).

| shape50 | humans$^\dagger$ | AlexNet | GoogLeNet | ResNet-50 | DCNNs$^\dagger$ |
|---|---|---|---|---|---|
| humans$^\dagger$ | 1 | | | | |
| AlexNet | .23 | 1 | | | |
| GoogLeNet | .07 | .09 | 1 | | |
| ResNet-50 | .13 | .02 | .33* | 1 | |
| DCNNs$^\dagger$ | .15 | .64** | .54** | .63** | 1 |
| **shape150** | | | | | |
| humans$^\dagger$ | 1 | | | | |
| AlexNet | .38* | 1 | | | |
| GoogLeNet | .25 | -.01 | 1 | | |
| ResNet-50 | .22 | -.02 | .29 | 1 | |
| DCNNs$^\dagger$ | .46* | .56** | .67* | .59** | 1 |
| **shape250** | | | | | |
| humans$^\dagger$ | 1 | | | | |
| AlexNet | .26 | 1 | | | |
| GoogLeNet | .30 | .08 | 1 | | |
| ResNet-50 | .07 | -.23 | .42* | 1 | |
| DCNNs$^\dagger$ | .29 | .49** | .70** | .61** | 1 |
| **color** | | | | | |
| humans$^\dagger$ | 1 | | | | |
| AlexNet | .25 | 1 | | | |
| GoogLeNet | -.04 | -.17 | 1 | | |
| ResNet-50 | -.06 | -.18 | .49** | 1 | |
| DCNNs$^\dagger$ | .14 | .45* | .61** | .69** | 1 |

## 3.4 Pretrained Deep Neural Networks

Further tests carried out on pretrained DCNNs without a fine-tuning process encourage the selected study design and point towards findings assumed but unproven in preceding analyses. As ResNet-50 (Mdn = 80.95) achieves performance indistinguishable from human observers (Mdn = 80.95) on shape50 examples (W = 636.00, z = -.71, p = .480, n = 63, r = -.09) without previous learning history of these distortions, generally accuracy of pretrained neural networks decreases drastically across levels of manipulation (see Figure 7). Pretrained neural networks frequently link consequences of applied distortion with image features learned from the data of 1000 classes and therefore fall for unintended classifications (e.g., shape manipulations as *snail* or *coil*, and green color manipulations as *cucumber*) due to the absence of category rules like *all images show 1 of 7 species of monkeys, apes, and lemurs*.

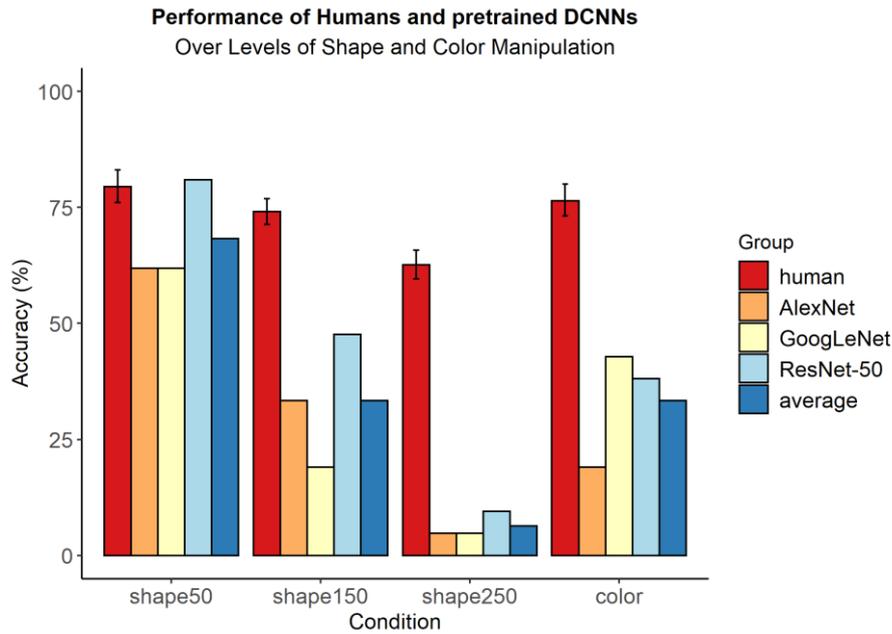

**Figure 7.** Performance of humans and pretrained DCNNs over levels of shape and color manipulation. Performance of pretrained DCNNs shows a sharp decline across manipulations, as networks start to confuse consequences of manipulation with object features due to missing limitation of classes. Error bars indicate 95 %-confidence intervals for human participants only, as DCNNs show fixed values without any variance.

# 4 Discussion

The presented attempt aims to compare performance of human participants' biological vision and feedforward neural networks' artificial vision in core object recognition of natural and manipulated images. Therefore, in the anticipated overarching hypotheses, selected DCNNs were thought to be a match for human performance in natural examples, and yet to be surpassed by their accuracy in novel manipulation images. Even though the expected results for control images from the ImageNet data set differ from numerous findings prevalent in the field of computer vision [1, 19, 57], as all three DCNNs fell short of achieving performance statistically comparable to human participants, additional analysis on pretrained models as well as prevalent underpowered training conditions hint at the existence of uncovered effects. Our results are consistent with most previous work [26, 27, 56] as image manipulation leads to performance decreases in both visual processing systems, while humans outperform DCNNs in all shape and color conditions by a significant margin. These findings fit with the assumption that humans share great robustness towards color variation as object categorization does not rely on color cues [50], while also being affected by shape distortion as it scales down both image quality and real-world fit.

That aside, correspondence between the brain and feedforward neural networks is still a matter of debate. Generally, as two different approaches ask different notions of this question, they may have to be answered in different ways. In the field of computational neuroscience, the question *Do they work in the same way?* has to be answered with *Not quite* for manipulations. Even though early visual processing in the brain seems to share vast similarities with feedforward neural networks in control images [42], numerous work has pointed out the importance of recurrent connectivity, as in ITC or vPFC, that seems to equip primates with the ability of core object recognition in challenge images. As neural networks experience severe problems with these *late-solved* images, the primate brain simply requires additional computation time in higher visual processing regions [43, 45]. More interestingly, a recent study by Kar and DiCarlo (2020) [58] verified this by proposing that drug-induced inactivation of these same prominently recurrent regions makes the primate brain behave more like a feedforward neural network. The list of processes involved in biological vision is long and exceeds limitations of neural networks modelling feedforward sweeps of signals as next to local and long-range feedforward, lateral recurrent, as well as local and long-range feedback connections are required [21]. It has to be noted that most neural network models are oversimplifications of the primate visual system as they start at the level of V1 and therefore neglect most precortical processing units such as RGC and LGN [59].

In the field of machine learning, the question *Do they perform on the same level?* concerning images including unexperienced manipulations would clearly have to be answered with *Not yet*. In this way, the existing performance differences postulated by previous studies for a wide spectrum of manipulations can be supported and extended by our findings. Theoretically, it could be argued that nonstop data collection in a continuous stream of perception and consciousness could give the brain an unfair advantage over artificial neural networks, which are just fed with a sparse selection of examples. While this definitely could be the case for color alterations, as for example human observers might have possibly seen a painted version of a blue gorilla at some point in their life, it is certainly impossible for not naturally occurring, unseen types of distortion. Therefore, we claim that effects observed for shape manipulations withstand this possible point of criticism similarly to novel eidolon distortion experiments [27, 56]. Another aspect that might seem difficult to manage in a behavioral comparison is the human ability to guess. In this particular case, the chosen study design eliminates this problem as DCNNs use a process similar to human guessing, when taking their best prediction of many calculated probabilities for a confined spectrum of categories.

For an interim conclusion, it can be summarized that in this approach we try to perform a fairer human-machine comparison by adopting selected measures proposed by Firestone (2020) [51]. Machines were limited like humans as DCNNs were only provided with the same small number of training and validation examples compared to human participants. While in contrast, humans were limited like machines using a rather specific *expert* data set, that humans most likely had to learn first, and the additional utilization of unexperienced manipulations without any evolutionary advantages as opposed to noise that might have been perceived by humans (e.g., in different weather conditions such as fog and snow) before. On top of that, human participants were only allowed to answer in a forced-choice format without an *other* category to reduce their responses to the capabilities of DCNNs.

However, it is certainly plausible that several limitations might have influenced the obtained results. First, as the study was designed as an online experiment, it is difficult to guarantee conduction under controlled conditions. Second, the tested data set contained only few images per category, which could have negatively impacted the fine-tuning process of DCNNs. Unfortunately, it is hard to reconstruct to what extent neural networks profited from features learned as pretrained models on great quantity of ImageNet data and how many features had to be newly acquired. Therefore, further data collection on a larger number of examples would be needed to determine exactly how image manipulation affects core visual object recognition under controlled conditions. On a final note, it must be mentioned that human object categorization involves many cognitive processes [60] which can be attributed to attention, memory, generalization, abstraction, and possibly even creativity.

In conclusion, the evidence from this work suggests that humans outperform feedforward neural networks in visual core object recognition of unexperienced shape and especially color manipulations on an innovative classification paradigm. These findings are consistent with a growing body of literature, that hints towards recurrence as a crucial factor for adequate generalization abilities. While this approach clearly has some limitations, we hope that its documentation might also serve as a springboard for the increasingly important discussion on fairness in human-machine comparisons.

# References

[1] Serre, T. (2019). Deep learning: The good, the bad, and the ugly. *Annu. Rev. Vis. Sci., 5*, 399-426. doi:10.1146/annurev-vision-091718-014951

[2] Biederman, I. (1987). Recognition-by-components: A theory of human image understanding. *Psychol. Rev., 94*, 115-147. doi:10.1037/0033-295X.94.2.115

[3] Pinto, N., Cox, D. D., and DiCarlo, J. J. (2008). Why is real-world visual object recognition hard? *PLoS Comput. Biol., 4*, 0151-0156. doi:10.1371/journal.pcbi.0040027

[4] Rajalingham, R., Issa, E. B., Bashivan, P., Kar, K., Schmidt, K., and DiCarlo, J. J. (2018). Large-scale, high-resolution comparison of the core visual object recognition behavior of humans, monkeys, and state-of-the-art deep artificial neural networks. *J. Neurosci., 38*, 7255-7269. doi:10.1523/jneurosci.0388-18.2018

[5] Cadieu, C. F., Hong, H., Yamins, D. L. K., Pinto, N., Ardila, D., Solomon, E. A., Majaj, N. J., and DiCarlo, J. J. (2014). Deep neural networks rival the representation of primate IT cortex for core visual object recognition. *PLoS Comput. Biol., 10*, e1003963. doi:10.1371/journal.pcbi.1003963

[6] Tanaka, K. (1996). Inferotemporal cortex and object vision. *Annu. Rev. Neurosci., 19*, 109-139. doi:10.1146/annurev.ne.19.030196.000545

[7] Riesenhuber, M., and Poggio, T. (1999). Hierarchical models of object recognition in cortex. *Nat. Neurosci., 2*, 1019-1025. doi:10.1038/14819

[8] Rolls, E. T. (2000). Functions of the primate temporal lobe cortical visual areas in invariant visual object and face recognition. *Neuron, 27*, 205-218. doi:10.1016/s0896-6273(00)00030-1

[9] DiCarlo, J. J., Zoccolan, D., and Rust, Nicole C. (2012). How does the brain solve visual object recognition? *Neuron, 73*, 415-434. doi:10.1016/j.neuron.2012.01.010

[10] Khaligh-Razavi, S.-M., and Kriegeskorte, N. (2014). Deep supervised, but not unsupervised, models may explain IT cortical representation. *PLoS Comput. Biol., 10*, e1003915. doi:10.1371/journal.pcbi.1003915

[11] Kriegeskorte, N. (2015). Deep neural networks: A new framework for modeling biological vision and brain information processing. *Annu. Rev. Vis. Sci., 1*, 417-446. doi:10.1146/annurev-vision-082114-035447

[12] Yamins, D. L., and DiCarlo, J. J. (2016). Using goal-driven deep learning models to understand sensory cortex. *Nat. Neurosci., 19*, 356-365. doi:10.1038/nn.4244

[13] Yamins, D. L., Hong, H., Cadieu, C., and DiCarlo, J. J. (2013). "Hierarchical modular optimization of convolutional networks achieves representations similar to macaque IT and human ventral stream", in: *Proceedings of the 26th International Conference on Neural Information Processing Systems (NeurIPS)*: Curran Associates Inc., 3093–3101.

[14] Geirhos, R., Jacobsen, J.-H., Michaelis, C., Zemel, R., Brendel, W., Bethge, M., and Wichmann, F. A. (2020). Shortcut learning in deep neural networks. *arXiv preprint arXiv:1312.6199*.

[15] Russakovsky, O., Deng, J., Su, H., Krause, J., Satheesh, S., Ma, S., Huang, Z., Karpathy, A., Khosla, A., Bernstein, M., Berg, A. C., and Fei-Fei, L. (2015). ImageNet Large Scale Visual Recognition Challenge. *Int. J. Comput. Vis., 115*, 211-252. doi:10.1007/s11263-015-0816-y

[16] Krizhevsky, A., Sutskever, I., and Hinton, G. E. (2012). "Imagenet classification with deep convolutional neural networks", in: *Adv. Neural Inf. Process. Syst.*, 1097-1105.

[17] Szegedy, C., Liu, W., Jia, Y., Sermanet, P., Reed, S., Anguelov, D., Erhan, D., Vanhoucke, V., and Rabinovich, A. (2015). "Going deeper with convolutions", in: *Proceedings of the IEEE Computer Society Conference on Computer Vision and Pattern Recognition (CVPR)*, 1-9.

[18] He, K., Zhang, X., Ren, S., and Sun, J. (2016). "Deep residual learning for image recognition", in: *Proceedings of the IEEE Computer Society Conference on Computer Vision and Pattern Recognition (CVPR)*, 770-778.

[19] LeCun, Y., Bengio, Y., and Hinton, G. (2015). Deep learning. *Nature, 521*, 436-444. doi:10.1038/nature14539

[20] Stabinger, S., Rodríguez-Sánchez, A., and Piater, J. (2016). "25 years of cnns: Can we compare to human abstraction capabilities?", in: *International Conference on Artificial Neural Networks (ICANN)*: Springer, 380-387.

[21] Kietzmann, T. C., McClure, P., and Kriegeskorte, N. (2018). Deep neural networks in computational neuroscience. *bioRxiv*, 133504. doi:10.1101/133504

[22] Cichy, R. M., and Kaiser, D. (2019). Deep neural networks as scientific models. *Trends Cogn. Sci., 23*, 305-317. doi:10.1016/j.tics.2019.01.009

[23] Ciresan, D., Meier, U., Masci, J., and Schmidhuber, J. (2012). Multi-column deep neural network for traffic sign classification. *Neural Networks, 32*, 333-338. doi:10.1016/j.neunet.2012.02.023

[24] Lee, K., Zung, J., Li, P., Jain, V., and Seung, H. S. (2017). Superhuman accuracy on the SNEMI3D connectomics challenge. *arXiv preprint arXiv:1706.00120*.

[25] Phillips, P. J., Yates, A. N., Hu, Y., Hahn, C. A., Noyes, E., Jackson, K., … O'Toole, A. J. (2018). Face recognition accuracy of forensic examiners, superrecognizers, and face recognition algorithms. *PNAS, 115*, 6171. doi:10.1073/pnas.1721355115

[26] Dodge, S., and Karam, L. (2017). "A study and comparison of human and deep learning recognition performance under visual distortions", in: *26th International Conference on Computer Communication and Networks (ICCCN)*: IEEE, 1-7.

[27] Geirhos, R., Temme, C. R., Rauber, J., Schütt, H. H., Bethge, M., and Wichmann, F. A. (2018b). "Generalisation in humans and deep neural networks", in: *Proceedings of the 32nd Conference on Neural Information Processing Systems (NIPS)*, 7538-7550.

[28] Hendrycks, D., Zhao, K., Basart, S., Steinhardt, J., and Song, D. (2019). Natural adversarial examples. *arXiv preprint arXiv:1907.07174*.

[29] Beery, S., Van Horn, G., and Perona, P. (2018). "Recognition in Terra Incognita", in: *Proceedings of the European Conference on Computer Vision (ECCV)*, 456-473.

[30] Rosenfeld, A., Zemel, R., and Tsotsos, J. K. (2018). The elephant in the room. *arXiv preprint arXiv:1808.03305*.

[31] Alcorn, M. A., Li, Q., Gong, Z., Wang, C., Mai, L., Ku, W.-S., and Nguyen, A. (2019). "Strike (with) a pose: Neural networks are easily fooled by strange poses of familiar objects", in: *Proceedings of the IEEE Computer Society Conference on Computer Vision and Pattern Recognition (CVPR)*, 4845-4854.

[32] Geirhos, R., Rubisch, P., Michaelis, C., Bethge, M., Wichmann, F. A., and Brendel, W. (2018a). ImageNet-trained CNNs are biased towards texture; increasing shape bias improves accuracy and robustness. *arXiv preprint arXiv:1811.12231*.

[33] Azulay, A., and Weiss, Y. (2019). Why do deep convolutional networks generalize so poorly to small image transformations? *J. Mach. Learn. Res., 20*, 1-25.

[34] Brendel, W., and Bethge, M. (2019). Approximating cnns with bag-of-local-features models works surprisingly well on imagenet. *arXiv preprint arXiv:1904.00760*.

[35] Jacobsen, J.-H., Behrmann, J., Zemel, R., and Bethge, M. (2018). Excessive invariance causes adversarial vulnerability. *arXiv preprint arXiv:1811.00401*.

[36] Nguyen, A., Yosinski, J., and Clune, J. (2015). "Deep neural networks are easily fooled: High confidence predictions for unrecognizable images", in: *Proceedings of the IEEE Computer Society Conference on Computer Vision and Pattern Recognition (CVPR)*, 427-436.

[37] Cichy, R. M., Khosla, A., Pantazis, D., Torralba, A., and Oliva, A. (2016). Comparison of deep neural networks to spatio-temporal cortical dynamics of human visual object recognition reveals hierarchical correspondence. *Sci. Rep., 6*, 27755. doi:10.1038/srep27755

[38] Bell, A. H., Hadj-Bouziane, F., Frihauf, J. B., Tootell, R. B. H., and Ungerleider, L. G. (2009). Object representations in the temporal cortex of monkeys and humans as revealed by functional magnetic resonance imaging. *J. Neurophysiol., 101*, 688-700. doi:10.1152/jn.90657.2008

[39] Connolly, A. C., Guntupalli, J. S., Gors, J., Hanke, M., Halchenko, Y. O., Wu, Y.-C., Abdi, H., and Haxby, J. V. (2012). The representation of biological classes in the human brain. *J. Neurosci., 32*, 2608. doi:10.1523/jneurosci.5547-11.2012

[40] Kiani, R., Esteky, H., Mirpour, K., and Tanaka, K. (2007). Object category structure in response patterns of neuronal population in monkey inferior temporal cortex. *J. Neurophysiol., 97*, 4296-4309. doi:10.1152/jn.00024.2007

[41] Nayebi, A., Bear, D., Kubilius, J., Kar, K., Ganguli, S., Sussillo, D., DiCarlo, J. J., and Yamins, D. L. (2018). "Task-driven convolutional recurrent models of the visual system", in: *Adv. Neural Inform. Proc. Sys. (NeurIPS)*, 5290-5301.

[42] Xu, Y., and Vaziri-Pashkam, M. (2020). Limited correspondence in visual representation between the human brain and convolutional neural networks. *bioRxiv*. doi:10.1101/2020.03.12.989376

[43] Kar, K., Kubilius, J., Schmidt, K., Issa, E. B., and DiCarlo, J. J. (2019). Evidence that recurrent circuits are critical to the ventral stream's execution of core object recognition behavior. *Nat. Neurosci., 22*, 974-983. doi:10.1038/s41593-019-0392-5

[44] Yamane, Y., Carlson, E. T., Bowman, K. C., Wang, Z., & Connor, C. E. (2008). A neural code for three-dimensional object shape in macaque inferotemporal cortex. *Nat. Neurosci., 11*(11), 1352-1360. doi:10.1038/nn.2202

[45] van Bergen, R. S., and Kriegeskorte, N. (2020). Going in circles is the way forward: the role of recurrence in visual inference. *arXiv preprint arXiv:2003.12128*.

[46] Douglas, R. J., & Martin, K. A. (2004). Neuronal circuits of the neocortex. *Annu. Rev. Neurosci., 27*, 419-451. doi:10.1146/annurev.neuro.27.070203.144152.

[47] Angelucci, A., and Bressloff, P. C. (2006). Contribution of feedforward, lateral and feedback connections to the classical receptive field center and extra-classical receptive field surround of primate V1 neurons. *Prog. Brain. Res., 154*, 93-120. doi:10.1016/S0079-6123(06)54005-1

[48] Kreiman, G., and Serre, T. (2020). Beyond the feedforward sweep: feedback computations in the visual cortex. *ANN NY Acad. Sci., 1464*, 222-241. doi:10.1111/nyas.14320

[49] Lamme, V. A., and Roelfsema, P. R. (2000). The distinct modes of vision offered by feedforward and recurrent processing. *Trends Neurosci., 23*, 571-579. doi:10.1016/s0166-2236(00)01657-x

[50] Delorme, A., Richard, G., and Fabre-Thorpe, M. (2000). Ultra-rapid categorisation of natural scenes does not rely on colour cues: a study in monkeys and humans. *Vis. Res., 40*, 2187-2200. doi:10.1016/S0042-6989(00)00083-3

[51] Firestone, C. (2020). Performance vs. competence in human–machine comparisons. *PNAS, 117*(43), 26562-26571. doi:10.1073/pnas.1905334117

[52] Scorolli, C., and Borghi, A. M. (2015). Square bananas, blue horses: the relative weight of shape and color in concept recognition and representation. *Front. Psychol., 6*, 1542. doi:10.3389/fpsyg.2015.01542

[53] Mathôt, S., Schreij, D., & Theeuwes, J. (2012). OpenSesame: An open-source, graphical experiment builder for the social sciences. *Behav. Res. Methods*, *44*(2), 314-324. doi:10.3758/s13428-011-0168-7

[54] Wickham, H. (2016). *ggplot2: Elegant Graphics for Data Analysis.* New York: Springer-Verlag.

[55] Pallant, J. (2007). *SPSS survival manual—A step by step guide to data analysis using SPSS for windows.* Maidenhead: Open University Press.

[56] Geirhos, R., Janssen, D. H., Schütt, H. H., Rauber, J., Bethge, M., and Wichmann, F. A. (2017). Comparing deep neural networks against humans: object recognition when the signal gets weaker. *arXiv preprint arXiv:1706.06969*.

[57] He, K., Zhang, X., Ren, S., and Sun, J. (2015). "Delving deep into rectifiers: Surpassing human-level performance on imagenet classification", in: *Proceedings of the IEEE International Conference on Computer Vision (ICCV)*, 1026-1034.

[58] Kar, K., and DiCarlo, J. J. (2020). Fast recurrent processing via ventral prefrontal cortex is needed by the primate ventral stream for robust core visual object recognition. *bioRxiv*. doi:10.1101/2020.05.10.086959

[59] Medathati, N. V. K., Neumann, H., Masson, G. S., and Kornprobst, P. (2016). Bio-inspired computer vision: Towards a synergistic approach of artificial and biological vision. *Comput. Vis. Image Und., 150*, 1-30. doi:10.1016/j.cviu.2016.04.009

[60] Pylyshyn, Z. (1999). Is vision continuous with cognition?: The case for cognitive impenetrability of visual perception. *Behav. Brain Sci., 22*, 341-365. doi:10.1017/S0140525X99002022